\documentclass{article}

\PassOptionsToPackage{numbers, compress}{natbib}

\usepackage[final]{neurips_2025}
\usepackage{tabularx} 

\usepackage{wrapfig}
\usepackage{calc}
\usepackage{multirow}
\usepackage{algpseudocode}
\usepackage[utf8]{inputenc} 
\usepackage[T1]{fontenc}    
\usepackage{url}            
\usepackage{booktabs}       
\usepackage{amsfonts}       
\usepackage{nicefrac}       
\usepackage{microtype}      
\usepackage{xcolor}         
\usepackage{xpatch}
\usepackage{graphicx}  
\usepackage{algorithm}
\usepackage{algorithmicx}
\usepackage{algpseudocode}
\usepackage{listings}
\usepackage{fancyvrb}
\usepackage{adjustbox}
\usepackage{natbib}
\usepackage{caption}
\usepackage{subcaption}
\usepackage{amsmath,amsfonts}
\usepackage{booktabs}
\usepackage{arydshln}
\usepackage{wrapfig}
\usepackage{xpatch}
\usepackage{hyperref}       
\makeatletter
\xapptocmd{\NAT@bibsetnum}{\setlength{\leftmargin}{0pt}\setlength{\itemindent}{\labelwidth}\addtolength{\itemindent}{\labelsep}}{}{}
\makeatother
\title{ZigzagPointMamba: Spatial-Semantic Mamba for Point Cloud Understanding}

\author{%
  Linshuang Diao$^{1,2}$ \quad 
  Sensen Song$^{1,2}$\thanks{Corresponding author} \quad 
  Yurong Qian$^{1,2}$ \quad 
  Dayong Ren$^{3}$\footnotemark[1] \\
  \\
$^1$Key Laboratory of Signal Detection and Processing, Xinjiang University \\
$^2$Joint International Research Laboratory of  \\
Silk Road Multilingual Cognitive Computing, Xinjiang University \\
$^3$National Key Laboratory for Novel Software Technology, \\
Nanjing University, Nanjing 210023, China \\
  \\
  \texttt{\ songsensen@stu.xju.edu.cn, rdyedu@gmail.com}
}
\begin{document}

\maketitle

\begin{abstract}
State Space models (SSMs) like PointMamba provide efficient feature extraction for point cloud self-supervised learning with linear complexity, surpassing Transformers in computational efficiency. However, existing PointMamba-based methods rely on complex token ordering and random masking, disrupting spatial continuity and local semantic correlations. We propose \textbf{ZigzagPointMamba} to address these challenges. The key to our approach is a simple zigzag scan path that globally sequences point cloud tokens, enhancing spatial continuity by preserving the proximity of spatially adjacent point tokens. Yet, random masking impairs local semantic modeling in self-supervised learning. To overcome this, we introduce a Semantic-Siamese Masking Strategy (SMS), which masks semantically similar tokens to facilitate reconstruction by integrating local features of original and similar tokens, thus overcoming dependence on isolated local features and enabling robust global semantic modeling. Our pre-training ZigzagPointMamba weights significantly boost downstream tasks, achieving a 1.59\% mIoU gain on ShapeNetPart for part segmentation, a 0.4\% higher accuracy on ModelNet40 for classification, and 0.19\%, 1.22\%, and 0.72\% higher accuracies respectively for the classification tasks on the OBJ-BG, OBJ-ONLY, and PB-T50-RS subsets of ScanObjectNN. Code is available at \textcolor{red}{https://github.com/Rabbitttttt218/ZigzagPointMamba}.
\end{abstract}
\section{Introduction}
Deep learning-based point cloud analysis requires models to extract and interpret intricate geometric features from unstructured spatial data. Traditional approaches, such as MLP\cite{qi2017pointnet,qi2017pointnet++} and Transformer architecture\cite{qi2023contrast,he2025contrastive}, are often hindered by high computational complexity, substantial resource demands, and limited generalization across diverse datasets. To address these challenges, PointMamba\cite{liang2024pointmamba} has emerged as a novel framework. By leveraging a state-space model, PointMamba achieves linear computational complexity and robust global feature aggregation, effectively balancing architectural simplicity with efficiency. Its strong knowledge transfer abilities, especially in self-supervised learning, have shown excellent performance and driven significant research into PointMamba-based models.

Despite the efficiency of PointMamba-based methods in global feature aggregation through state-space models, their dependence on conventional scanning schemes—such as random, Hilbert, or Z-order approaches—often disrupts spatial continuity, leading to suboptimal performance in downstream tasks like point cloud segmentation and classification. For instance, random scanning fragments local geometric coherence, while Hilbert and Z-order schemes struggle to adapt to complex topologies, resulting in disjointed token sequences that impair feature consistency. To address this, we propose ZigzagPointMamba, a novel method that employs a zigzag scanning path to globally sequence point cloud tokens while preserving spatial proximity. By generating smoother, spatially coherent token sequences, ZigzagPointMamba enhances the quality of feature representations, with preliminary experiments demonstrating a 1.59\% mIoU improvement in part segmentation on ShapeNetPart and a 0.4\% accuracy gain in classification on ModelNet40.

\begin{figure}[t!]
    \centering
    \begin{subfigure}{0.44\textwidth}  
        \centering
        \includegraphics[height=6cm]{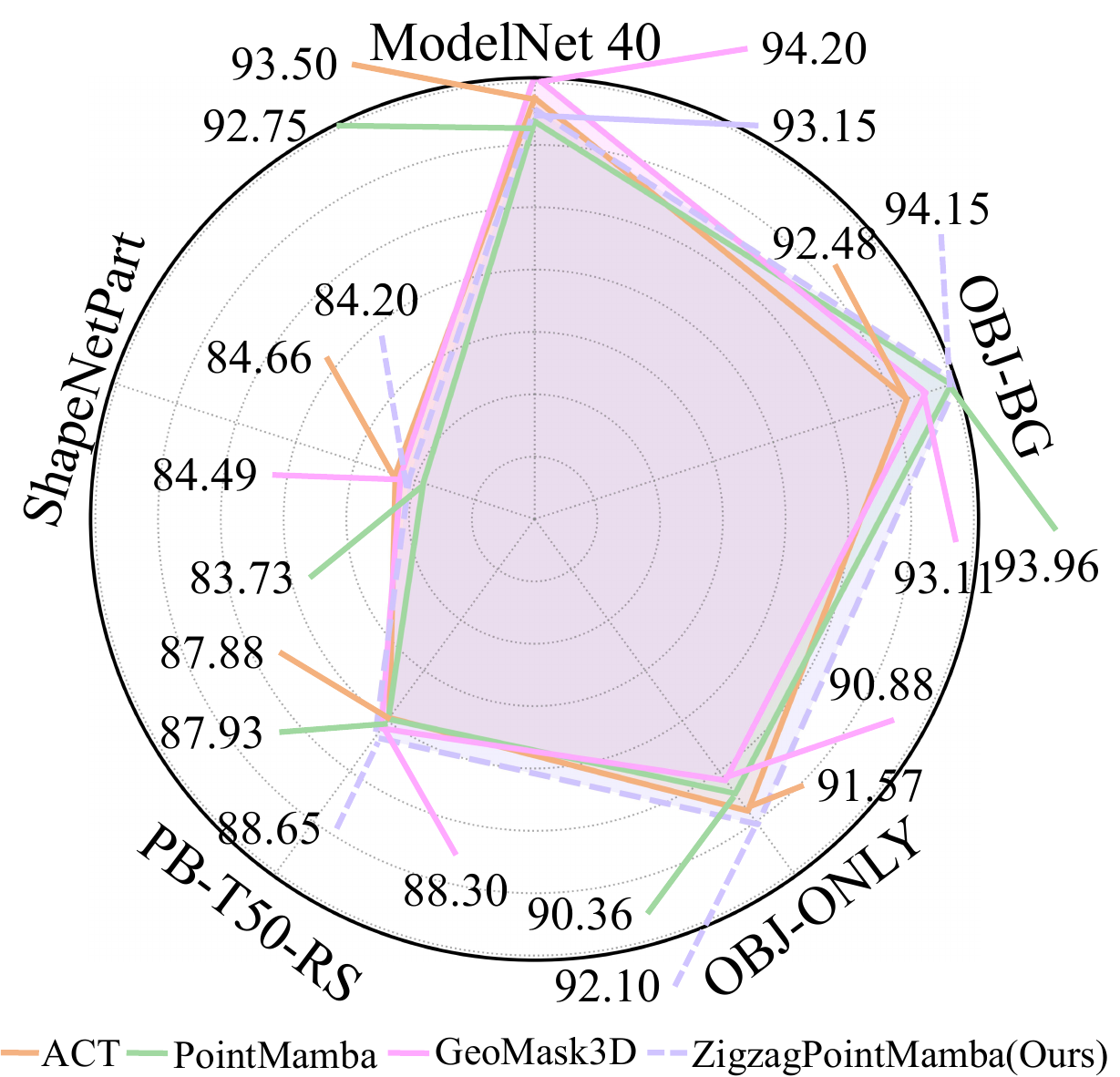}  
        \caption{Comparision of Performance}
        \label{(1.a)}
    \end{subfigure}\hspace{0pt}
    \hfill
    \begin{subfigure}{0.325\textwidth}  
        \centering
        \includegraphics[height=5.6cm]{Reconstructionofchair.pdf}  
        \caption{Comparison of the Effects of SMS and Random Masking}
        \label{(1.b)}
    \end{subfigure}\hspace{0pt}
    \hfill
    \begin{subfigure}{0.21\textwidth}  
        \centering
        \includegraphics[height=6cm]{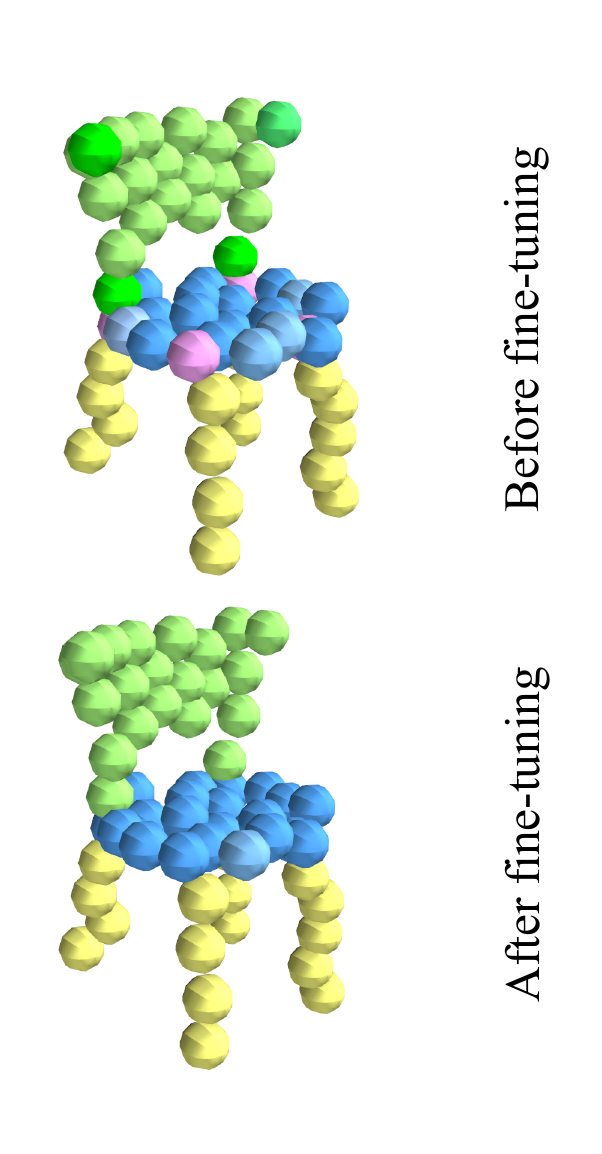}  
        \caption{Features Before and After Fine-tuning}
        \label{1.c}
    \end{subfigure}
    \caption{As can be seen from Fig. \ref{1} (a), compared with ACT, PointMamba, and GeoMask3D, our proposed ZigzagPointMamba performs better on the ScanObjectNN dataset. Fig. \ref{1} (b) presents a stark contrast between the effects of SMS and random masking, highlighting the superiority of our proposed method in terms of reconstruction. Fig. \ref{1} (c) demonstrates the features before and after fine-tuning, indicating the effectiveness of our method in refining feature representations.}
    \label{1}
\end{figure}

Building on the spatially coherent token sequences provided by ZigzagPointMamba, we further enhance PointMamba’s self-supervised learning capabilities by addressing limitations in its masking strategy. Traditional random masking, which relies on adjacent tokens to reconstruct masked regions, struggles to capture global semantic dependencies, compromising performance in downstream point cloud tasks such as segmentation and classification. To overcome this, we introduce the Semantic-Siamese Masking Strategy (SMS), which leverages the smooth token sequences from ZigzagPointMamba to mask semantically similar local structures, thereby strengthening token-level semantic associations. By integrating local features of original and semantically related tokens, SMS enables robust global semantic modeling, yielding more accurate feature representations. Preliminary results validate the synergy of ZigzagPointMamba and SMS, achieving up to 1.22\% accuracy improvements across classification on ScanObjectNN subsets (OBJ-BG, OBJ-ONLY, PB-T50-RS).

SMS employs a Siamese-like comparison to evaluate semantic similarity between point cloud tokens, enabling targeted masking of coherent structures, such as entire object parts like a chair’s armrest or a car’s wheel, rather than random token selections. By leveraging smooth, spatially continuous tokens, SMS ensures that masked semantic tokens maintain both spatial proximity and semantic continuity, preserving the topological integrity of the point cloud during self-supervised learning. Specifically, SMS operates through two key steps: (1) The point cloud is decomposed into fine-grained semantic tokens by leveraging the zigzag scanning order, and (2) A Siamese-like mechanism evaluates token-wise similarity scores, and tokens exceeding a predefined similarity threshold are masked. This strategy eliminates redundant local features, forcing the model to reconstruct masked regions by relying on global semantic context from retained tokens. This strategy mitigates the limitations of traditional random masking, which relies solely on local information and disrupts semantic coherence, thereby optimizing global feature modeling and enhancing self-supervised learning capabilities. Combined with ZigzagPointMamba’s spatial advantages, SMS achieves superior reconstruction quality and more distinct feature distributions after fine-tuning, Our \textbf{ZigzagPointMamba} achieves excellent performance on various point cloud analysis datasets (as shown in Fig \ref{1} (a).). In addition, the reconstruction effect of SMS we proposed is better, and the feature distribution after fine-tuning is also more distinct (as shown in Fig.\ref{1} (b) and Fig.\ref{1} (c)). 

Collectively, our contributions include: (1) a zigzag scan path that preserves spatial proximity, mitigating discontinuities in traditional scanning methods; (2) the SMS approach, which enhances token-level semantic associations for robust global feature modeling; and (3) the integration of spatial and semantic continuity in ZigzagPointMamba, significantly advancing point cloud analysis, particularly in self-supervised learning applications.

\section{Related Works}
\subsection{Point Cloud Analysis Methods Based on MLP }
Early deep learning methods for point cloud processing\cite{ren2024mffnet,guo2024lidar,ren2022point}, such as PointNet\cite{qi2017pointnet}, used shared multi-layer perceptrons (MLPs) for feature extraction and max pooling to aggregate global information. However, these methods had limitations in modeling local geometric structures. PointNet++\cite{qi2017pointnet++} improved this by introducing hierarchical MLPs and farthest point sampling (FPS) for multi-scale feature learning. Subsequently, methods like RandLA-Net\cite{hu2020randla} proposed lightweight architectures that processed large-scale point clouds through random sampling and local feature aggregation mechanisms, significantly enhancing computational efficiency. PointMLP\cite{ma2022rethinking} constructed networks purely based on residual MLPs and introduced a local geometric affine module, achieving certain results in point cloud tasks and providing a network architecture foundation for the application of subsequent self-supervised learning methods in point cloud processing. Later, \cite{zheng2024sa} enhanced the MLP architecture through efficient addition and shift operations, reducing computational complexity while improving point cloud classification performance. HPE\cite{zou2024improved} improved MLPs using high-dimensional positional encoding, mapping the 3D coordinates of points to a high-dimensional space to effectively enhance the representation of point cloud positional information and further boost the MLP's ability to model local structures. PointMT\cite{zheng2024pointmt} combines the efficiency of MLPs with the global feature capture capability of Transformers, and introduces innovative mechanisms to address the computational bottlenecks of traditional Transformers, enabling efficient point cloud analysis. 

\subsection{The Deep Evolution of Transformer Architecture and Masking Strategies}
In recent years, Transformer-based point cloud processing methods have made synergistic progress in both architectural design and masking strategies. PCT\cite{guo2021pct} first introduced self-attention mechanisms into point cloud processing, while Point Transformer V1\cite{zhao2021point} enhanced geometric modeling through vector attention. Subsequently, Point Transformer V2\cite{wu2022point} and OctFormer\cite{wang2023octformer} improved computational efficiency via hierarchical attention and octree partitioning, respectively. GPSFormer\cite{wang2024gpsformer} and DAPoinTr\cite{li2025dapointr} have achieved innovative integration of graph structures and dynamic path mechanisms. while Siamese-based approaches\cite{liu2025revisiting} have demonstrated versatile Transformer architectures for 3D tracking tasks. Meanwhile, the evolution of masking strategies is deeply intertwined with Transformer architecture development: Point-BERT\cite{yu2022point} was the first to adapt the masked pre-training paradigm from natural language processing to point clouds, establishing a self-supervised learning framework through coordinate reconstruction tasks. Recent advances in self-supervised learning\cite{wu2023self} have further explored contrastive learning approaches for point cloud understanding. Point-MAE\cite{pang2022masked} further strengthened feature learning by increasing the masking ratio, while Point-M2AE\cite{zhang2022point} introduced a multi-scale masking framework that synergizes with Transformer's hierarchical architecture. Methods such as GeoMAE\cite{tian2023geomae} and SemMAE\cite{li2022semmae} incorporate geometric\cite{chen2024geosegnet} awareness and semantic guidance into masking design, enabling models to better understand spatial layouts while maintaining computational efficiency.
\subsection{The Development of State-Space Models in Point Cloud Processing }
Due to the quadratic complexity of Transformers, State Space Models (SSMs) have become a research hotspot for their linear computational complexity\cite{ren2024spiking}. The Mamba model\cite{gu2023mamba} demonstrated the high-efficiency potential of SSMs in sequence processing and proposed an effective linear-time sequence modeling method. In further research, the attention mechanism of the Mamba model was explored, revealing how to enhance the capture of key information while maintaining efficient computation\cite{ali2024hidden}. Subsequently, many Mamba-based variants have been applied to various fields\cite{liu2024vmamba,wang2024mamba,wang2024mamba}. In the point cloud domain, recent developments include spectral-informed approaches\cite{bahri2025spectral} for robust processing and voxel-based methods\cite{zhang2024voxel} for 3D object detection. For example, U-Mamba\cite{ma2024u} enhances long-range dependency modeling in biomedical image segmentation, and LocalMamba\cite{huang2024localmamba} further optimizes the performance of visual State Space Models by introducing windowed selective scanning. However, these models still face significant challenges when processing complex geometric structure data such as point clouds.

PointMamba\cite{liang2024pointmamba} was the first to apply the SSM framework to point cloud processing, achieving remarkable progress in feature extraction. However, its scan method may hinder the capture of geometric structures, while random masking may obscure key information. To address these issues, we propose the ZigzagPointMamba model. By combining the zigzag scan path and the Semantic-Siamese Masking Strategy (SMS), ZigzagPointMamba ensures that spatially adjacent points maintain proximity and effectively learns global semantics.
\section{Methods}
\subsection{Background: State Space Models and PointMamba}
State Space Models (SSMs) serve as a classical framework for processing sequential and spatial data,capturing long-range dependencies through a recursive state transition equation:
\begin{equation}
h_t = A h_{t-1} + B x_t,
\label{eq:basic_ssm}\tag{1}
\end{equation}
where $h_t$ is the state at time step $t$, $x_t$ is the observation (or input) at step $t$, and $A$, $B$ are learned transition matrices. This model exhibits $O(n)$ linear computational complexity, making it highly efficient for large-scale data modeling. 

In the field of point cloud analysis, PointMamba adapts the SSM framework to process unordered point clouds by treating each point as an independent state unit. The modified state transition is:
\begin{equation}
h_t = A_t h_{t-1} + B_t x_t,\tag{2}
\label{eq:pointmamba}
\end{equation}
where $h_t \in \mathbb{R}^d$ represents the state of point,$x_t \in \mathbb{R}^m$ contains the point's 3D coordinates and features, $A_t, B_t$ are dynamic parameters conditioned on $x_t$.

\subsection{The structure of ZigzagPointMamba}

We present \textbf{ZigzagPointMamba}, an enhanced architecture derived from PointMamba that introduces two key innovations: (1) a novel \textbf{zigzag scan path} for structured point cloud traversal, and (2) a \textbf{Semantic-Siamese Masking Strategy (SMS)} for enhanced token-level semantic associations for robust global feature modeling.

\begin{figure}
    \centering
    \includegraphics[width=1\linewidth]{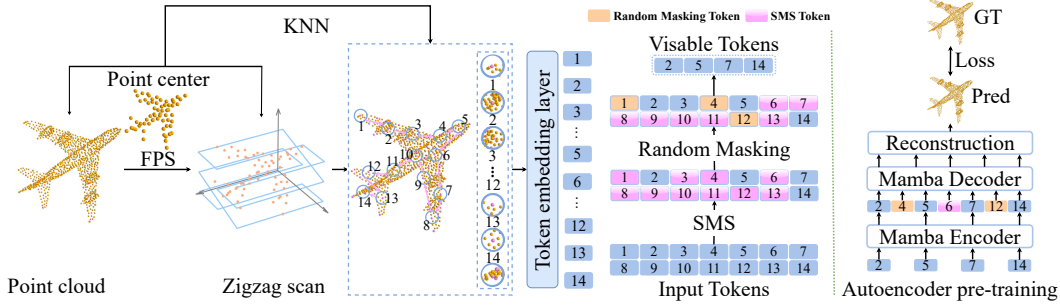}
\vspace{-10pt} 
    \caption{ZigzagPointMamba pre-training pipeline.\textbf{ }Select key point cloud points with FPS. Extract feature labels via KNN algorithm and lightweight PointNet. Serialize using the zigzag scan path. Input serialized features into a point cloud MAE architecture with SMS for training, obtaining point cloud feature representations and providing parameters for downstream tasks.}
            \label{fig:2}
\end{figure}

In the pre-training process of the ZigzagPointMamba architecture (Fig. \ref{fig:2}), input point cloud data first undergoes the zigzag scan path. Key points are selected via Farthest Point Sampling (FPS), followed by zigzag scan applied on XY, XZ, and YZ planes. Hierarchical layering and alternating sorting generate traversal paths with clear spatial logic, transforming unordered point clouds into ordered sequences to enhance feature spatial proximity and continuity. As illustrated, feature vectors of adjacent points in the ordered sequence better reflect their 3D adjacency, establishing a foundation for capturing local geometric features. Subsequently, features are extracted from these paths using KNN and a lightweight PointNet to generate point tokens. The SMS calculates semantic similarity from token feature vectors, identifies redundant regions via thresholding, and applies masking to prevent the model from over-relying on local information, thereby enhancing the capture of long-range semantic relationships.

To process 3D point cloud data and assist deep learning models in capturing spatial structure, we use an improved zigzag scan path approach. This approach generates structured traversal paths on orthogonal planes in 3D space. In traditional 2D image processing, the 2D zigzag scan path\cite{hu2024zigma} is a commonly used method. As shown in the left sub-figure of Fig. \ref{3}, the 2D zigzag scan path traverses data points in a specific zigzag pattern, effectively organizing the data order in 2D space. However, when dealing with 3D point cloud data, the scenario becomes more complex. A 3D point cloud can be represented as \(P=\{p_i \}_{i = 1}^N \), where \(p_i = (x_i, y_i, z_i) \in \mathbb{R}^3 \). Our 3D zigzag scan path extends the 2D approach by performing scanning operations on three key planes: the XY-plane, XZ-plane, and YZ-plane, to comprehensively analyze the spatial relationships within 3D point clouds. In implementation, the point cloud is first divided into layers along different coordinate axes through coordinate-based layering; subsequently, segment-based alternating sorting is used to construct traversal scan paths with specific zigzag patterns on each of the three key planes. The effect is illustrated in the right sub-figure of Fig. \ref{3}. 
\subparagraph{Scan generation on the XY-Plane} First, we sort all points in ascending order based on the Z-coordinate. This step arranges the point cloud according to the values of Z, which facilitates the subsequent layering operation. After sorting, the sorted point cloud is divided into \(L_{xy} \) layers, where \(L_{xy} = \left\lceil \frac{M}{3} \right\rceil \). For the \(k\)-th layer, denoted as \(L_k^{xy} \), it is represented as:
\[
L_k^{xy} = \left\{p_i \in P \mid z_{(k-1)\frac{N}{L_{xy}}} \leq z_i < z_{k\frac{N}{L_{xy}}} \right\}, \quad k = 1, 2, \dots, L_{xy}.\tag{3}
\]
Here, \(z_{(k-1)\frac{N}{L_{xy}}} \) and \(z_{k\frac{N}{L_{xy}}} \) are the values of the Z-coordinates at the corresponding positions in the sorted point cloud. By this layering, the point cloud is divided along the Z-direction, ensuring that points within the same layer are similar in their Z-coordinates.

Within each layer \(L_k^{xy} \), the points are first sorted in ascending order based on their X-coordinates. This operation organizes the points along the X-direction within each layer. Next, the points are divided into \(S\) segments, where \( S = \min\left(\left\lfloor \frac{|L_k^{xy}|}{d} \right\rfloor, m \right) \), ensuring that each segment contains approximately \(d \) points. For the \( \mathbf{s}\)-th segment \(S_s^{xy} \), the points are alternately sorted by the Y-coordinate:
When \( s \) is even, the points are sorted in ascending order of Y. That is, for \( p_i, p_j \in S_s^{xy} \), if \(y_i < y_j \), then \(p_i \) comes before \(p_j \).
When \( s \) is odd, the points are sorted in descending order of Y. That is, for \( p_i, p_j \in S_s^{xy} \), if \(y > y_i \), then \(p_i \) comes before \(p_j \). This alternating sorting creates a zigzag scan path within each layer. Finally, the segments are connected in sequence to form a scan \(R_k^{xy} \).

\subparagraph{Scan generation on the XZ-Plane} 
Similar to the XY-plane, points are sorted ascendingly by the Y-coordinate first. The point cloud is divided into \(L_{xz} \) layers, where \(L_{xz} = \left\lfloor \frac{M}{3} \right\rfloor + I_{M \bmod 3 \geq 1} \). For the \( k \)-th layer \(L_k^{xz} \), it is defined as \( L_k^{xz} = \left\{p_i \in P \mid y_{(k - 1)\frac{N}{L_{xz}}} \leq y_i < y_{k\frac{N}{L_{xz}}} \right\}, \, k = 1, \cdots, L_{xz} \). Points in each layer are sorted by X-coordinate, segmented, and sorted by Z-coordinate to form \(R_k^{xz} \).

\subparagraph{Scan generation on the YZ-Plane} 
Likewise, points are sorted by the X-coordinate first. The point cloud is divided into \(L_{yz} \) layers, where \(L_{yz} = \left\lfloor \frac{M}{3} \right\rfloor \). For the \(k\)-th layer \( L_k^{yz} \), it is defined as \( L_k^{yz} = \left\{p_i \in P \mid x_{(k - 1)\frac{N}{L_{yz}}} \leq x_i < x_{k\frac{N}{L_{yz}}} \right\}, \,k = 1, \cdots, L_{yz} \). After sorting points in each layer by Y-coordinate, segmenting, and sorting by Z-coordinate, we get \(R_k^{yz} \).

\newcounter{savefig}
\setcounter{savefig}{\value{figure}}
\vspace{-0.8cm}

\begin{minipage}[t]{0.48\textwidth}
  \begin{figure}[H] 
    \centering
    \includegraphics[width=\linewidth]{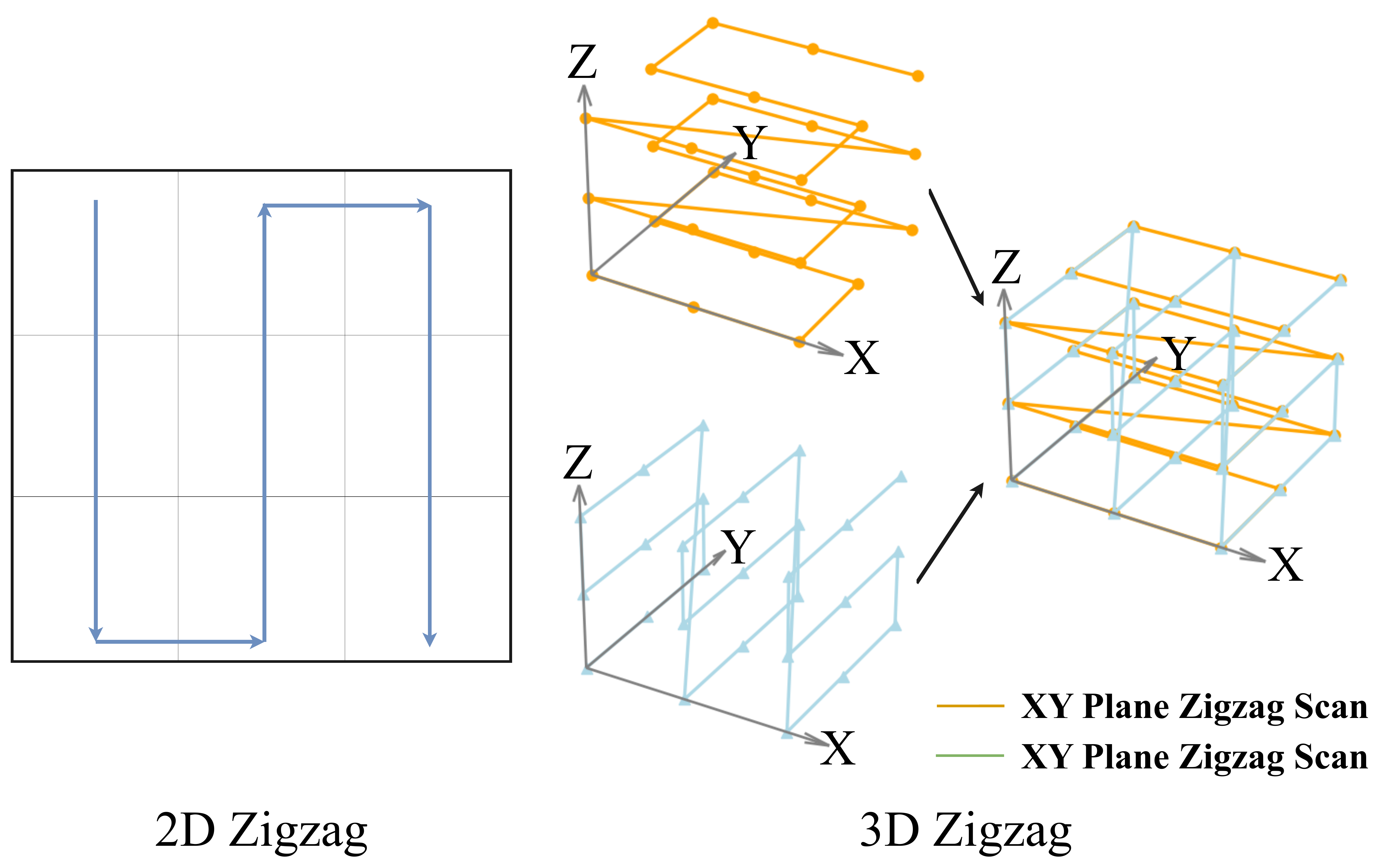}
    \caption{Comparison of 2D and 3D zigzag. The 3D strategy scans on multiple planes. As an extension of the 2D one, it aids the model in preserving spatial proximity.}
    \label{3}
  \end{figure}
\end{minipage}%
\hfill%
\begin{minipage}[t]{0.48\textwidth}
  \begin{figure}[H]
    \centering
    \includegraphics[width=\linewidth]{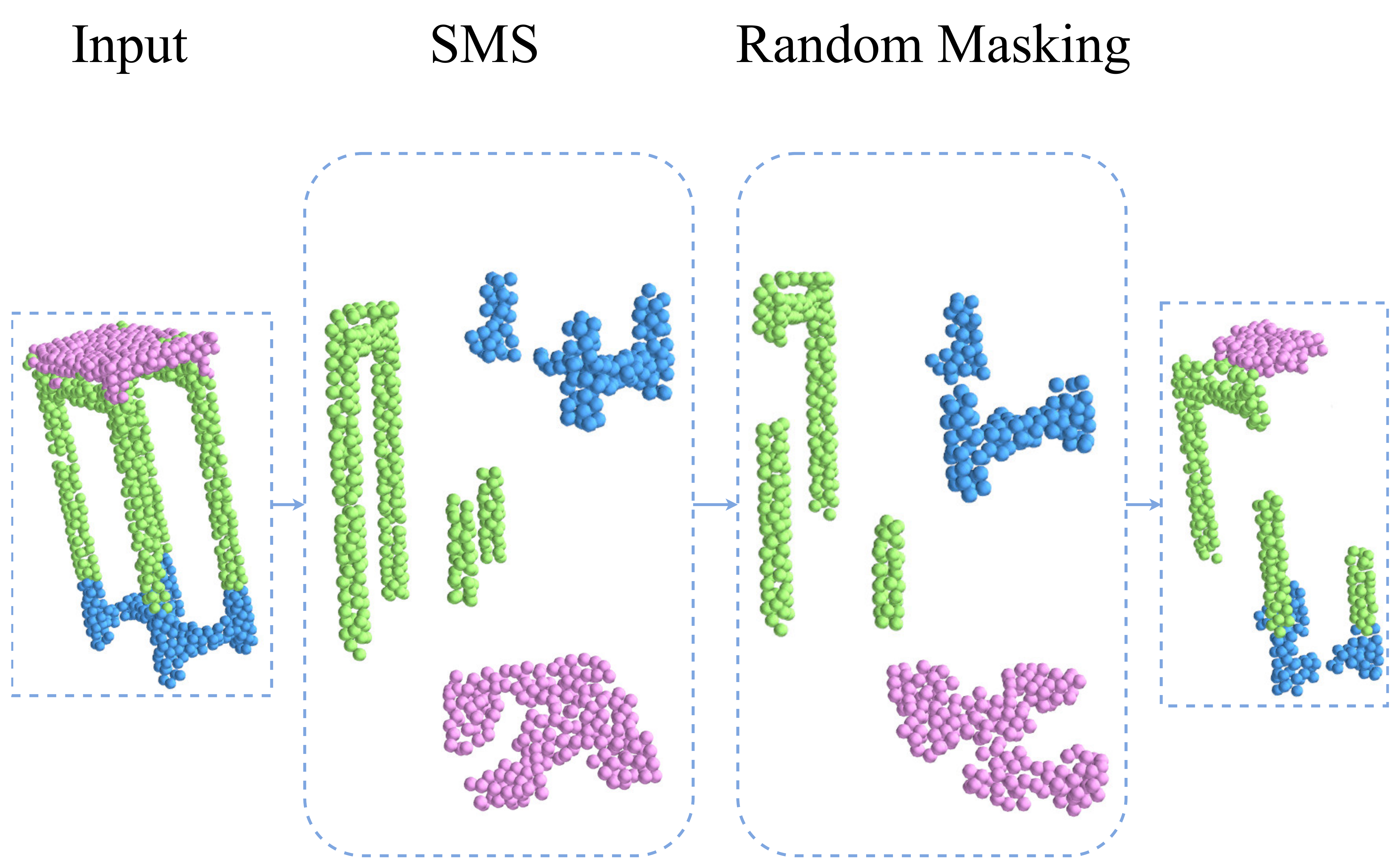}
    \caption{Details of Masking. Leverage SMS to mask out tokens with high semantic feature similarity in the point cloud. Then, apply random masking to a subset of the remaining tokens to enhance the robustness of the pre-training model.}
    \label{4}
  \end{figure}
\end{minipage}
\setcounter{figure}{\value{savefig}}
\addtocounter{figure}{2} 
\subsection{Masking Process}\label{masking-process}
The traditional random masking strategy is still prone to losing key information and disrupting the semantic coherence and geometric structural relationships that have been sorted out when masking the point cloud data processed by the zigzag scan path. Although the point cloud tokens arranged by the zigzag scan path ensure the continuity of spatially adjacent features, the random masking fails to make good use of this advantage. To address this issue, we introduce the SMS. Based on the semantic similarity of point cloud features, the SMS selects the masking regions. By fully leveraging the structured foundation established by the zigzag scan, it can not only preserve the important semantic information and geometric structure but also enhance the model's ability to model global semantic features, significantly improving the performance of the MAE in handling downstream tasks.

\subsubsection{Semantic-Siamese Masking Strategy (SMS)}
\begin{algorithm}
  \caption{Semantic-Siames Masking Strategy}
  \label{alg:1}
  \begin{algorithmic}
  \item[] \textbf{Input:} \(group\_input\_tokens\): point cloud feature tensor with shape \(B, G, C \) (batch\_size, number of groups, feature\_dimension).\\
\hspace{3em}\(threshold\): SMS retention \(threshold\) (default 0.8), controlling the proportion of tokens to retain.
  \item[] \textbf{Output:} \(bool\_masked\_pos\): boolean mask tensor with shape \(B, G, C \), indicating which tokens are masked.\\
  \item[1:] \(B, G, C \leftarrow \)shape(\(group\_input\_tokens\)) // Get tensor dimensions
  \item[2:] \(tokens\_norm \leftarrow \text{F.normalize}(group\_input\_tokens, \text{dim}=-1)\) // Normalize feature vectors to unit length
  \item[3:] \(similarity\_matrix \leftarrow \text{torch.bmm}(tokens\_norm, tokens\_norm^T).\text{clamp}(0, 1)\) // Compute cosine similarity matrix and clamp to [0,1]
  \item[4:] \(redundancy\_score \leftarrow \sum_{\text{dim}=-1}(similarity\_matrix)\) // Calculate redundancy score for each token
  \item[5:] \(k \leftarrow \max(1, \lfloor threshold \times G \rfloor)\) // Determine number of tokens to retain (at least 1)
\item[6:] \textbf{if} \(k = 0\)
\\
\qquad \textbf{return} \(\textnormal{torch.zeros}([B, G], \textnormal{dtype}=\textnormal{torch.bool})\) 

  \item[7:] \(thresholds \leftarrow \text{torch.topk}(redundancy\_score, k=k, \text{largest}=\text{torch.False}).values[:, -1]\) // Get k-th smallest redundancy score as threshold
  \item[8:] \(bool\_masked\_pos \leftarrow redundancy\_score > thresholds\) // Generate mask (tokens with higher redundancy are masked)
  \item[9:] \textbf{return} \(bool\_masked\_pos\)
  \end{algorithmic}
  \end{algorithm}

The main goal of SMS is to remove the parts of the input data that are irrelevant or redundant to the task, keeping the most representative and informative tokens (As shown in Algorithm\ref{alg:1}, its effect is presented in Fig. \ref{1} (b) and Fig. \ref{4}). This process takes place during the self-supervised learning phase of ZigzagPointMamba.First, for the input \(T\) (with shape \( B \times G \times C \)), we normalize the tokens to ensure that the features of each token are transformed into unit vectors, thus eliminating scale differences between different feature dimensions. The specific formula is:
\[
T_{norm_{b,g,c}} = \frac{T_{b,g,c}}{\sqrt{\sum_{c'=0}^{C - 1} T_{b,g,c'}^2}},\tag{4}
\]
where \( b \in \{1, \cdots, B\} \), \( g \in \{1, \cdots, G\} \), and \( c \in \{1, \cdots, C\} \). This step converts the feature vector of each token into a unit vector, ensuring that the features have a uniform scale.

Next, we compute the cosine similarity between the normalized feature vectors, resulting in a similarity matrix. The similarity between two tokens \(i\) and \(j\) in batch \(b\) is calculated using the following formula:
\[
S_{b,i,j} = \frac{f_{b,i} \cdot f_{b,j}}{\|f_{b,i}\| \|f_{b,j}\|},\tag{5}
\]
where \(f_{b,i}\) and \(f_{b,j}\) are the normalized feature vectors of the tokens \(i\)-th and \(j\)-th in batch \(b\)-th respectively. Since we have already normalized the feature vectors, \(\|f_{b,i}\| = 1\) and \(\|f_{b,j}\| = 1\),so the formula simplifies to:
\[
S_{b,i,j} = \sum_{c = 0}^{C - 1} T_{norm_{b,i,c}} \cdot T_{norm_{b,j,c}}.\tag{6}
\]
Also, we limit the values of the similarity matrix to the range \([0, 1]\) via the following transformation:\(S_{b,i,j} = \max\left(0, \min\left(1, S_{b,i,j}\right)\right).\)
This matrix represents the similarity between every pair of tokens, where \( i\) and \( j\) are indices of two different tokens. Then, we sum each row of the similarity matrix to obtain the redundancy score for each token, as shown in the following formula:
\[
R_{b,i} = \sum_{j = 0}^{G - 1}S_{b,i,j}.\tag{7}
\]
This redundancy score reflects how similar each token is to the other tokens. A higher redundancy score indicates that the token is more redundant. Based on the semantic threshold \(t_{semantic}\), we calculate the number of tokens that need to be retained, denoted as \(k = \lfloor t_{semantic} \cdot G \rfloor.\)Next, we sort the redundancy scores and choose the \( k \)-th smallest redundancy score as the threshold for each batch:
\[
t_b = \text{topk}(R_b, k = k, largest = \text{False})[k - 1].\tag{8}
\]
This threshold is used to identify redundant tokens, which are marked for masking, generating the initial semantic mask. The formula for mask generation is:
\[
M_{semantic_{b,i}} = 
\begin{cases} 
\text{True}, & \text{if } R_{b,i} > t_b, \\
\text{False}, & \text{otherwise}    .\tag{9}
\end{cases}
\]
Tokens with redundancy scores exceeding the threshold are marked as True for masking.

\subsubsection{Random Mask Generation}
After generating the SMS, we proceed with random masking of the remaining tokens. The main purpose of this stage is to further reduce redundancy by introducing randomness and to avoid overfitting of the model. When masking is required, we randomly select a certain percentage of tokens from those that have not been masked by the SMS. The number of tokens to be masked is calculated using the following formula:\(N_{mask_b} = \lfloor R_{mask} \cdot |I_{available_b}| \rfloor.\)Then, we randomly choose the indices of these available tokens and mark them as masked:\(M_{final_{b,I_{mask_b}}} = \text{True}.\)

\section{Experiments}
We present in this section the experimental configuration, datasets, and evaluation results for assessing the performance of \textbf{ZigzagPointMamba}. Our experimental framework incorporates the newly proposed zigzag scan path and SMS. Extensive evaluation across multiple benchmark datasets on various downstream tasks demonstrates that our approach achieves superior performance improvements over existing methods in terms of classification accuracy, segmentation performance, and model robustness.

\subsection{Pre-training Setup}

In the pre-training phase \cite{yu2022point,zheng2024point} , self-supervised learning extracts generalizable representations from unannotated point cloud data, providing transferable parameters for downstream tasks. We introduce the zigzag scan path and SMS to boost the model's local structure capture. To adapt to different resolutions, point clouds are patch-divided linearly (1024 points into 64 patches with 32 points per patch via KNN). The encoder has 12 vanilla Mamba blocks (384 dims each), and the decoder uses 4 Mamba blocks for reconstruction. We randomly select one path, setting the random masking ratio at 0.6 and SMS ratio at 0.8. Training uses AdamW optimizer (lr = 0.001), Cosine annealing scheduler, 300 epochs, batch size 128, and loss of Chamfer distance L2 \cite{bakshi2023near}. Experiments are run on a NVIDIA A40 GPU.  

\begin{table}[t!]
\centering
\captionsetup{font=footnotesize, labelfont=bf} 
\vspace{-10.5pt}  
\caption{\textbf{Object Classification on ScanObjectNN Dataset.} We conducted experiments on three subsets of the ScanObjectNN dataset: the OBJ-BG subset, OBJ-ONLY subset, and PB-T50-RS subset.}
\renewcommand{\arraystretch}{1.1} 
\scriptsize 
\begin{tabular*}{\linewidth}{@{}l@{\extracolsep{\fill}}cccccc@{}}
    \specialrule{1.12pt}{0pt}{0pt} 
    Methods       & Reference        & Param.(M)     & FLOPs(G)     & OBJ-BG    & OBJ-ONLY   & PB-T50-RS \\
    \specialrule{0.7pt}{0pt}{0pt} 
    Point-Bert\cite{yu2022point}  & CVPR 22  & 22.1          & 4.8          & 87.3      & 88.12      & 83.07     \\
    MaskPoint\cite{liu2022masked}  & CVPR 22  & 22.1          & 4.8          & 89.70     & 89.30      & 84.60     \\
    PointMAE\cite{pang2022masked}  & ECCV 22  & 22.1          & 4.8          & 90.02     & 88.29      & 84.60     \\
    PointM2AE\cite{zhang2022point}   & NeurIPS 22 & 15.3          & 3.6          & 91.22     & 88.81      & 86.43     \\
    ACT\cite{dong2022autoencoders}   & ICLR 23 & 22.1          & 4.8          & 93.29     & 91.91      & 88.21     \\
    ReCon\cite{qi2023contrast}   & ICML 23 & 43.6          & 5.3          & \textbf{94.15}& \textbf{93.12}& \textbf{89.73}\\ 
    GeoMask3D\cite{bahri2024geomask3d}   & TMLR 25 & -          & -          & 93.11     & 90.36      & 88.30     \\
    \specialrule{0.7pt}{0pt}{0pt} 
    PointMamba(\cite{liang2024pointmamba})(baseline)  & NeurIPS 24  & \textbf{12.3} & \textbf{3.1} &93.96           &90.88            &87.93           \\
    \textbf{ZigzagPointMamba(Ours)}  &   & \textbf{12.3} & \textbf{3.1} & \textbf{94.15} & 92.10 & 88.65 \\ 
    \specialrule{1.12pt}{0pt}{0pt} 
\end{tabular*}
\label{tab:1}
\end{table}
\begin{table}[htbp]
\centering
\captionsetup{font=small,labelfont=bf}
\vspace{-10.5pt}
\begin{minipage}[t]{0.49\textwidth} 
\centering
\small
\caption{\textbf{Classification on ModelNet40 Dataset.} We report the overall accuracy from 1024 points without voting.}
\resizebox{\linewidth}{!}{
\renewcommand{\arraystretch}{1.3}
\begin{tabular}{@{}lcccc@{}}
\specialrule{1.5pt}{0pt}{0pt}
Methods             &Reference & Param.(M)     & FLOPs(G)     & OA(\%) \\
\specialrule{1pt}{0pt}{0pt}
Point-Bert\cite{yu2022point} &CVPR 22 & 22.1 & 4.8 & 92.7 \\
MaskPoint\cite{liu2022masked} &CVPR 22 & 22.1 & 4.8 & 92.6 \\
PointMAE\cite{pang2022masked} &ECCV 22 & 22.1 & 4.8 & 93.2 \\
PointM2AE\cite{zhang2022point} &NeurIPS 22 & 15.3 & 3.6 & 93.4 \\
ACT\cite{dong2022autoencoders} &TCLR 23 & 22.1 & 4.8 & 93.6 \\
GeoMask3D\cite{bahri2024geomask3d} &TMLR 25 & - & - & \textbf{94.20} \\
\specialrule{1pt}{0pt}{0pt}
PointMamba(\cite{liang2024pointmamba})(baseline) & NeurIPS 24& 12.3 & 1.5 & 92.75 \\
\textbf{ZigzagPointMamba(Ours)}     & & \textbf{12.3} & \textbf{1.5} & 93.15 \\
\specialrule{1.5pt}{0pt}{0pt}
\end{tabular}}
\vspace{-3pt}
\label{tab:2}
\end{minipage}
\hfill
\begin{minipage}[t]{0.49\textwidth} 
\centering
\small
\caption{\textbf{Part Segmentation on ShapeNetPart Dataset.} The mIoU of all classes (Cls.) and instances (Inst.) is reported.}
\resizebox{\linewidth}{!}{
\renewcommand{\arraystretch}{1.113}
\begin{tabular}{@{}lccc@{}}
\specialrule{1.35pt}{0pt}{0pt}
Methods                        & Reference & Inst.mIoU & Cls.mIoU \\
\specialrule{0.8pt}{0pt}{0pt}
Point-BERT \cite{yu2022point}  & TMLR 25   & 85.6      & 84.1 \\
MaskPoint\cite{liu2022masked}  & TMLR 25   & 86.0      & 84.4 \\
PointMAE\cite{pang2022masked}  & ECCV 22   & 86.1      & 84.1 \\
PointM2AE\cite{zhang2022point} & NeurIPS 22& \textbf{86.51} & \textbf{84.86} \\
ACT\cite{dong2022autoencoders} & ICLR 23   & 86.14     & 84.66 \\
GeoMask3D\cite{bahri2024geomask3d} & TMLR 25 & 86.04   & 84.49 \\
\specialrule{0.8pt}{0pt}{0pt}
PointMamba(\cite{liang2024pointmamba})(baseline) & NeurIPS 24 & 85.28 & 82.57 \\
\textbf{ZigzagPointMamba(Ours)} &           & 85.78     & 84.16 \\
\specialrule{1.35pt}{0pt}{0pt}
\end{tabular}}
\vspace{-3pt}
\begin{flushleft}
\scriptsize
Our method uses 17.36M parameters and 5.5G FLOPs.
\end{flushleft}
\label{tab:3}
\end{minipage}
\end{table}
\vspace{-10pt}
\subsection{Datasets and Performance Evaluation of Downstream Tasks}
\begin{wraptable}{r}{0.5\textwidth}
    \centering
    \captionsetup{font=small,labelfont=bf}
    \normalsize  
    \caption{\textbf{Few-shot learning on ModelNet40}. A dedicated dataset for few-shot learning constructed based on ModelNet40.}
    \vspace{-0.5em}
    \resizebox{\linewidth}{!}{%
    \renewcommand{\arraystretch}{1.6}
    \begin{tabular}{@{}lcccccc@{}} 
        \specialrule{2pt}{0pt}{0pt}
        \multirow{2}{*}{Methods} & \multirow{2}{*}{Reference} & \multicolumn{2}{c}{5-way} & \multicolumn{2}{c}{10-way} \\
        \cline{3-6} 
        & & 10-shot & 20-shot & 10-shot & 20-shot \\
        \specialrule{1.25pt}{0pt}{0pt}
        Point-Bert\cite{yu2022point} & CVPR 22 & 94.6±3.0 & 96.3±2.5 & 91.0±5.0 & 92.7±4.8 \\
        MaskPoint\cite{liu2022masked} & CVPR 22 & 95.0±3.7 & 97.2±1.5 & 91.4±4.5 & 93.4±3.2 \\
        PointMAE\cite{pang2022masked} & ECCV 22 & 96.3±3.1 & 97.8±1.8 & 92.6±4.0 & 95.0±2.8 \\
        PointM2AE\cite{zhang2022point} & NeurIPS 22 & 96.8±2.0 & 98.3±1.5 & 92.3±4.2 & 95.2±2.5 \\
        ACT\cite{dong2022autoencoders} & ICLR 23 & 96.8±2.1 & 98.0±1.5 & 93.3±4.0 & 95.6±3.0 \\
        PointGPT-S\cite{chen2023pointgpt} & NeurIPS 23 & 96.8±1.8 & 98.6±1.2 & 92.6±3.5 & 95.2±2.5 \\
        ReCon\cite{qi2023contrast} & ICML 23 & \textbf{97.3±1.8} & 98.0±1.5 & \textbf{93.3±4.3} & \textbf{95.8±2.8} \\
        \specialrule{1.25pt}{0pt}{0pt}
        PointMamba(\cite{liang2024pointmamba})(baseline) & NeurIPS 24  & 96.0±2.0 & \textbf{99.0±1.0} & 88.5±2.4 & 93.8±1.2 \\
        \textbf{ZigzagPointMamba(Ours)} &  & 96.0±2.1 & \textbf{99.0±1.2} & 90.0±2.2 & 94.2±1.0  \\
        \specialrule{2pt}{0pt}{0pt}
    \end{tabular}}
    \label{tab:4}
\end{wraptable}
\textbf{ModelNet40}: 
In the classification experiment on ModelNet40~\cite{sun2022benchmarking}. As shown in Table  \ref{tab:2}, we conducted experiments without using the voting strategy. The classification accuracy of ZigzagPointMamba reached 93.15\%, representing an 0.4\% increase compared to PointMamba (Please note that the data of PointMamba in the experimental tables are obtained from our own experiments). The ModelNetFewShot dataset, constructed based on ModelNet40, is specifically designed for few-shot learning research. The selection of few-shot data in it strictly follows the following rules: randomly select 5 or 10 categories from the 40 categories of ModelNet40 as the task categories, and then randomly select 10 or 20 samples from each selected category as the support set. We conducted 15 independent few-shot learning experiments on ZigzagPointMamba using the ModelNetFewShot dataset. In the 10-way 10-shot scenario, the average classification accuracy of ZigzagPointMamba increased by 1.5\% compared to PointMamba; in the 10-way 20-shot scenario, the average accuracy increased by 0.4\%. As shown in Table~\ref{tab:4}, ZigzagPointMamba also demonstrates promising classification performance when handling few-shot data.

\textbf{ScanObjectNN}: 
As summarized in Table~\ref{tab:1}, we comprehensively evaluate \textbf{ZigzagPointMamba} across three distinct experimental configurations of ScanObjectNN~\cite{uy2019revisiting}:
OBJ-BG (objects with background), OBJ-ONLY (isolated objects), PB-T50-RS(perturbed objects with 50\% translation and random scaling). ZigzagPointMamba demonstrates consistent improvements over the baseline PointMamba in all scenarios: In OBJ-ONLY, ZigzagPointMamba achieves $\mathbf{92.10\%}$ classification accuracy ($\uparrow\mathbf{1.22\%}$ versus PointMamba's $90.88\%$), highlighting its efficacy in handling clean object geometries. For OBJ-BG with background interference, our method attains $\mathbf{94.15\%}$ accuracy ($\uparrow\mathbf{0.19\%}$ over PointMamba's $93.96\%$), showcasing robustness to environmental noise. Under the most challenging PB-T50-RS conditions, ZigzagPointMamba maintains $\mathbf{88.65\%}$ accuracy ($\uparrow\mathbf{0.72\%}$ versus $87.93\%$), validating its resilience to severe geometric perturbations.

\textbf{ShapeNetPart}: 
As can be seen from Table \ref{tab:3}, ZigzagPointMamba performs remarkably well in the segmentation task of ShapeNetPart\cite{chang2015shapenet}. Its Inst.mIoU reaches 85.78\%, which is 0.5\% higher than that of PointMamba. The Cls.mIoU is 84.16\%, 1.59\% higher than that of PointMamba.(The comparison of its segmentation effects is shown in the supplementary materials.)

\subsection{Analysis and ablation study}
To deeply explore the specific contributions of the zigzag scan path and SMS to the model performance, we conducted a series of ablation experiments on the OBJ-ONLY and PB-T50-RS datasets and observed the changes in the model performance.

\textbf{Influence of Different SMS Thresholds}: In the "Setting" column of the Table \ref{tab:7}, the paired values represent the random masking ratio and the SMS masking ratio respectively. Among them, when the SMS masking ratio is 0.8 and the random masking ratio is 0.6, the performance is optimal. The accuracies in the OBJ-ONLY and PB-T50-RS settings reach 92.08\% and 88.65\% respectively. When the threshold is lower than 0.8, redundant information will interfere with the model's learning process, affecting the accuracy of classification and segmentation. For example, when processing samples from the ScanObjectNN dataset, the model is easily interfered by noise features. When the threshold is higher than 0.8, over-masking occurs, resulting in the loss of key information. This makes it difficult for the model to grasp the features and semantic relationships in complex point cloud data, ultimately leading to a decline in performance.
\captionsetup{labelfont=bf} 
\begin{wraptable}{r}{0.55\textwidth} 
    \centering
    \footnotesize 
    \caption{The effect of different scanning curves.}
    \renewcommand{\arraystretch}{1.3} 
    \begin{tabular*}{\linewidth}{@{}l@{\extracolsep{\fill}}cc@{}}
        \specialrule{1.2pt}{0pt}{0pt}
        Scanning curve & \multicolumn{1}{c}{OBJ-ONLY} & \multicolumn{1}{c}{PB-T58-RS} \\
        \specialrule{0.7pt}{0pt}{0pt}
        Random & 92.60 & 90.18 \\
        Z-order and Trans-Z-order & \textbf{93.29} & 90.36 \\
        Hilbert and Z-order & \textbf{93.29} & 90.88 \\
        Trans-Hilert and Trans-Z-order & \textbf{93.29} & \textbf{91.91} \\
        \specialrule{0.7pt}{0pt}{0pt}
        Hilbert and Trans-Hilbert & 90.88 & 87.93 \\
        \textbf{zigzag scan path (Ours)} & 92.10 & 88.65 \\
        \specialrule{1.2pt}{0pt}{0pt}
    \end{tabular*}
    \label{tab:5}
\end{wraptable}

\textbf{Influence of Different Attention Mechanisms}: By comparing different attention mechanisms, we found that when using SMS, the model achieved accuracies of 92.08\% and 88.51\% in the OBJ-ONLY and PB-T50-RS settings, respectively, which are better than the standard attention mechanism and the multi-attention mechanism, as shown in Table \ref{tab:6}. This indicates that SMS can more effectively help the model capture semantic features and strengthen the modeling of long-range semantic relationships. Compared with the standard attention mechanism, SMS performs masking operations based on semantic similarity, which can more accurately screen out the information valuable for the model's learning and reduce the interference of redundant information, thus improving the model's performance in complex tasks. Although the multi-attention mechanism can pay attention to data from multiple perspectives, it is less effective than SMS in focusing on semantic features, resulting in a slightly inferior overall performance.
\begin{table}[htbp]
    \centering
    \captionsetup{font=small, labelfont=bf} 
    \footnotesize  
    \begin{minipage}[c]{0.49\textwidth}
        \centering
        \caption{The effect of the thresholds of different SMS.}
        \renewcommand{\arraystretch}{1.1}  
        \begin{tabularx}{\linewidth}{@{}l>{\centering\arraybackslash}X>{\centering\arraybackslash}X@{}}
            \specialrule{1.2pt}{2pt}{2pt}
            \multirow{2}{*}{Setting} & \multicolumn{2}{c}{OA(\%)} \\
            \cline{2-3} 
            & \multicolumn{1}{c}{\vspace{0.01em}OBJ-ONLY} & \multicolumn{1}{c}{\vspace{0.01em}PB-T50-RS} \\
            \specialrule{0.7pt}{2pt}{2pt}
            0.5 & 90.71 & 87.99 \\
            0.6 & 91.57 & 87.68 \\
            0.7 & 91.22 & 88.45 \\
            0.8 & \textbf{92.08} & \textbf{88.65} \\ 
            0.9 & 91.91 & 88.36 \\
            \specialrule{1.2pt}{2pt}{2pt}
        \end{tabularx}
        \label{tab:6}
    \end{minipage}
    \hfill 
    \begin{minipage}[c]{0.49\textwidth}
        \centering
        \caption{The effect of different attention.}
        \renewcommand{\arraystretch}{1.1}  
        \begin{tabularx}{\linewidth}{@{}l>{\centering\arraybackslash}X>{\centering\arraybackslash}X@{}}
            \specialrule{1.2pt}{2pt}{2pt}
            \multirow{2}{*}{Setting} & \multicolumn{2}{c}{OA(\%)} \\
            \cline{2-3} 
            & \multicolumn{1}{c}{OBJ-ONLY} & \multicolumn{1}{c}{PB-T58-RS} \\
            \specialrule{0.7pt}{2pt}{2pt}
            Attention & 91.22 & 88.17 \\
            Multi-Attention & 90.53 & 87.79 \\
            \textbf{SMS} & \textbf{92.08} & \textbf{88.51} \\ 
            \specialrule{1.2pt}{2pt}{2pt}
        \end{tabularx}
        \label{tab:7}
    \end{minipage}
\end{table}

\textbf{Influence of Different Scanning Curves}: We conducted an in-depth exploration of the impact of different scanning curves on the model performance. As shown in Table \ref{tab:5}, the zigzag scan path achieved accuracies of 92.10\% and 88.65\% in the OBJ-ONLY and PB-T50-RS settings, respectively. Compared with other scanning curve settings, the effect of this result is not particularly outstanding from a data perspective. However, we cannot solely evaluate the pros and cons of the zigzag scan path based on the increase in accuracy. In our actual model, the zigzag scan path is closely integrated with the SMS. The two complement each other and work together to improve the model performance. 

\subsection{Limitations and Future Work}
While ZigzagPointMamba demonstrates strong performance across various benchmarks, several limitations warrant acknowledgment. First, our SMS strategy employs a fixed similarity threshold ($\tau=0.8$), which may not generalize optimally across all point cloud distributions. Future work will explore adaptive thresholding mechanisms that dynamically adjust based on local semantic complexity. Second, the unidirectional modeling nature of state space models presents challenges when extending to temporal point cloud sequences or video-based applications, where simultaneous multi-directional context may be beneficial. Addressing these limitations through dynamic threshold adaptation and exploring bidirectional or temporal extensions of the zigzag scanning strategy represents promising directions for future research.

\section{Conclusion}
In this paper, we introduced \textbf{ZigzagPointMamba}, an innovative state-space model that addresses critical limitations in existing PointMamba-based approaches for point cloud self-supervised learning.  Through extensive experiments on multiple benchmark datasets, we demonstrated that our approach improves classification accuracy, segmentation performance, and robustness, especially under noisy and occluded conditions. Our results show that the zigzag scan path preserves spatial continuity in point clouds, while the SMS helps the model focus on global structures, preventing over-reliance on local features. Overall, ZigzagPointMamba provides a powerful pre-trained backbone that effectively supports downstream point cloud analysis tasks, offering a practical and robust foundation for a wide range of applications.

\clearpage 
\bibliographystyle{plain}
\bibliography{Reference.bib}

\end{document}